\documentclass[nohyperref]{article}

\usepackage{enumitem}
\usepackage{romannum}

\usepackage{adjustbox}
\usepackage{multirow}
\usepackage{tablefootnote}
\usepackage{threeparttable}

\usepackage{microtype}
\usepackage{graphicx}
\usepackage{subfigure}
\usepackage{booktabs} 

\usepackage{hyperref}

\usepackage[accepted]{main}

\usepackage{amsmath}
\usepackage{amssymb}
\usepackage{amsthm}
\usepackage{bm}
\usepackage{mathtools}

\usepackage[capitalize,noabbrev]{cleveref}

\theoremstyle{plain}

\theoremstyle{definition}

\theoremstyle{remark}

\usepackage[textsize=tiny]{todonotes}

\icmltitlerunning{Temporal Dynamic Embedding for Irregularly Sampled Time Series}

\begin{document}

\twocolumn[
\icmltitle{Temporal Dynamic Embedding for Irregularly Sampled Time Series}


\icmlsetsymbol{equal}{*}

\begin{icmlauthorlist}
\icmlauthor{Mincheol Kim}{dh,iphc}
\icmlauthor{Soo-Yong Shin}{dh,iphc,smc}
\end{icmlauthorlist}

\icmlaffiliation{dh}{Department of Digital Health, Samsung Advanced Institute for Health Sciences \& Technology, Sungkyunkwan University, Seoul 06355, Korea}
\icmlaffiliation{iphc}{Department of Intelligent Precision Healthcare Convergence, Sungkyunkwan University, Suwon 16419, Korea}
\icmlaffiliation{smc}{Center for Research Resource Standardization, Samsung Medical Center, Seoul 06351, Korea}

\icmlcorrespondingauthor{Soo-Yong Shin}{sy.shin@skku.edu}

\icmlkeywords{Irregularly Sampled Time Series, Missing value, Representation Learning, Healthcare}

\vskip 0.3in
]



\printAffiliationsAndNotice{}  

\begin{abstract}
In several practical applications, particularly healthcare, clinical data of each patient is 
individually recorded in a database at irregular intervals as required. This causes a sparse 
and irregularly sampled time series, which makes it difficult to handle as a structured 
representation of the prerequisites of neural network models. We therefore propose temporal 
dynamic embedding (TDE), which enables neural network models to receive data that change the 
number of variables over time. TDE regards each time series variable as an embedding vector 
evolving over time, instead of a conventional fixed structured representation, which causes a 
critical missing problem. For each time step, TDE allows for the selective adoption and 
aggregation of only observed variable subsets and represents the current status of patient 
based on current observations. The experiment was conducted on three clinical datasets: 
PhysioNet 2012, MIMIC-\Romannum{3}, and PhysioNet 2019. The TDE model performed competitively 
or better than the imputation-based baseline and several recent state-of-the-art methods with 
reduced training runtime.
\end{abstract}

\section{Introduction}
\label{sec:intro}

Early deep learning models for time series analysis such as recurrent neural networks (RNNs), 
have been developed only for well-structured data, which means regularly aligned sampled, 
fully observed, fixed-size multivariate time series \cite{fawaz2019}. With the outstanding 
performance of RNNs, time series observations are dominantly forced to be formed structured 
representation, that is, tensor composed with "batch size $\times$ sequence length $\times$ 
feature size". However, in many practical applications, such as healthcare \cite{yadav2018}, 
finance \cite{sezer2020}, meteorology \cite{ravuri2021}, and traffic engineering \cite{zhang2017},
each measurement is individually recorded in a database at irregular intervals and different 
unaligned time points caused by the incidents of each measurement. Furthermore, a measurement 
itself can contain meaningful information, along with absence or presence per se \cite{che2018}. 
These practical situations cause a multivariate sparse and irregularly sampled time series, 
which makes it difficult to handle a fixed dimensional structured representation.

Regardless of these critical constraints, many previous studies considered irregularly sampled 
time series as structured representations \cite{shukla2021survey}. Owing to the derived sparse 
missing problem, they established an imputation framework conditioned on observations to set 
up a structured representation. However, with the fixed dimensional structured representation, 
the input tensor should cover all variable sets for every time step, whereas the observed 
variables at each time step are much less than the entire variable set, which is commonly 
approximately to 10\% subsets in practical applications, which means a 90\% missing ratio 
\cite{che2018}. Even in some cases, some variables have never been measured for all time 
steps. Thus, it is a difficult issue that imputation is exactly accomplished and generalized 
or verified on its feasibility, owing to no true observations for missing. Additionally, the 
imputation framework did not consider inherent information for the absence or presence of the 
measurement itself.

In this study, we approach this challenge with a new perspective, which is rather than trying 
to establish full-imputed structured data, considering only observed data at each time point 
without any further inference such as imputation, which has the possibility of obtaining 
false information. We call this method temporal dynamic embedding (TDE). The TDE handles 
multivariate time series using a dynamic embedding technique. Dynamic embedding changes its 
representation over time, depending on the observed value. With irregularly sampled time 
series, each variable is individually represented to latent space, and the state at each 
time point is indicated by aggregating only a subset of observed variables and measurements 
and not the entire variable set. This approach is free from restrictions with imputation 
feasibility and quality; thus, it is robust because the TDE model makes a decision based 
on truly observed information.

Our main contributions are summarized as follows:
\begin{enumerate}[label=\arabic*)]
    \item We suggest a novel approach that allows the handling of irregularly sampled time 
    series itself without any data manipulation.
    \item We establish a temporal dynamic structure that can capture the representation of 
    each time status by selectively aggregating few observations at each time point.
    \item The TDE model shows state-of-the-art performance on three clinical tasks compared 
    to previous imputation-based methods.
\end{enumerate}

\section{Related Work}
\label{sec:related_work}

Previous studies on classification tasks for irregularly sampled time series can be broadly 
divided into two approaches: conventional structured representation and end-to-end learning. 
Previous research related to and motivated to the TDE model can be categorized as embedding 
representation learning.

\subsection{Conventional structured representation}
A basic approach for irregularly sampled time series is fixed temporal discretization to 
establish a structured representation \cite{marlin2012,lipton2016}. By discretizing 
continuous time series into hourly (or other time-scale) bins, it can be easily converted 
to a regularly aligned multivariate time series, which is followed by confronting a 
challenge with an empty bin, more than two observations in a bin, or blurring the exact 
measured time. Generally, the empty bin is considered as a missing point, although it 
takes a significantly large proportion of the dataset. Accordingly, many prior studies 
have been based on statistical imputation methods such as mean imputation, autoregressive 
integrated moving average (ARIMA) \cite{ansley1984}, multiple imputation by chained 
equations (MICE) \cite{van2011}, matrix factorization (MF) \cite{koren2009}, K-nearest 
neighborhood (KNN) \cite{friedman2001}, and expectation-maximization (EM) algorithm 
\cite{garcia-laencina2010}. Even zero imputation can be considered as one of the simplest 
ways because it has almost the same effect as dropout regularization \cite{srivastava2014}, 
except for normalization, and prevents the models from propagating missing inputs 
\cite{yi2020}. Furthermore, several recent studies have approached RNN-based imputation by 
establishing variants of RNNs \cite{che2018,cao2018,yoon2019,suo2020}. With an assumption, 
particularly in healthcare, that each variable tends to return to the normal range over 
time, \citeauthor{che2018} \citeyearpar{che2018} established a decay term on a gated 
recurrent unit (GRU) \cite{cho2014} that the last observed value decays to the mean of 
the variable over time and the hidden state also decays over time. \citeauthor{cao2018} 
\citeyearpar{cao2018} presented a generalized version of RNN-based imputation beyond a 
decay assumption by considering past observed values and the correlation of other 
variables at the measured time. Similarly, generative adversarial network (GAN) 
\cite{goodfellow2014}-based imputation methods were presented 
\cite{yoon2018,li2019,luo2019}. While these approaches achieve better imputation 
performance, they still rely on a fixed temporal discretization constraint, which is an 
unnecessary and risky assumption on downstream tasks with irregularly sampled time series. 
Receiving a sparse matrix as it is for inputs can be harmful to the stability of neural 
networks \cite{yi2020} because network propagation proportionally increases depending on 
the missing ratio. Additionally, if there is an error for imputation, its incorrect 
information will also explode along with the propagation.

\subsection{End-to-end learning}
Rather than explicit imputation over fixed temporal bins, Shukla \& Marlin 
\citeyearpar{shukla2019,shukla2021} established an end-to-end classification technique 
based on interpolation from irregularly sampled time series. \citeauthor{shukla2019} 
\citeyearpar{shukla2019} suggested several semi-parametric radial basis function 
(RBF)-based interpolation layers, which interpolate irregularly sampled time series 
into a set of reference time points. \citeauthor{shukla2021} \citeyearpar{shukla2021} 
proposed a multi-time attention module that also interpolates into a set of reference 
time points. However, unlike previous studies that simply performed classification tasks 
directly through interpolation, classification is performed through latent state via 
variational autoencoder (VAE) \cite{kingma2014}. Similarly, \citeauthor{horn2020} 
\citeyearpar{horn2020} suggested end-to-end classification for irregularly sampled 
time series by embedding an unordered set encoding and attention-based aggregation.

\begin{figure*}[ht]
    \vskip 0.2in
    \begin{center}
        \centerline{\includegraphics[width=\linewidth]{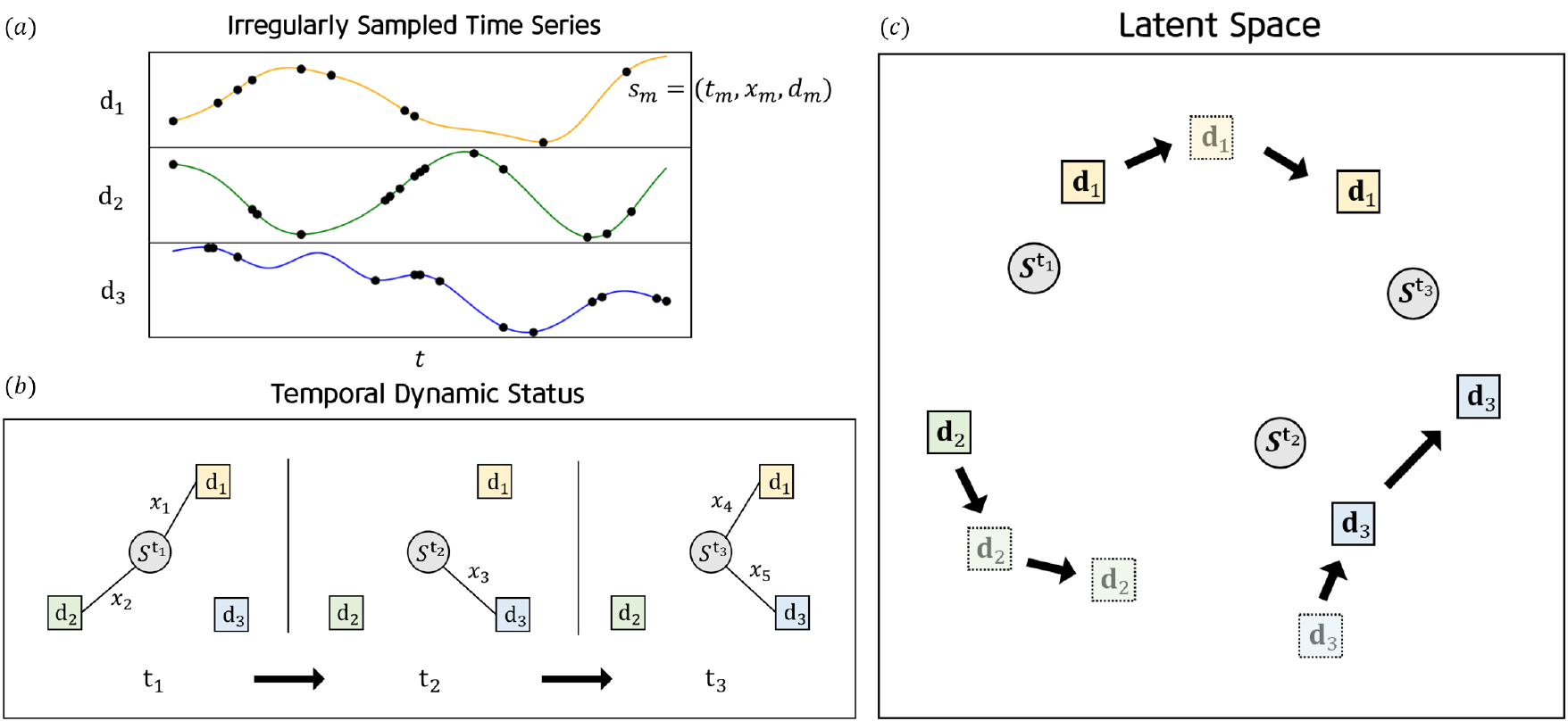}}
        \caption{\textbf{Overview of temporal dynamic embedding} 
        (a) Example of irregularly sampled multivariate time series, which consisted 
        of 3 variables (i.e., $d_m\in\{\mathrm{d}_1,\mathrm{d}_2,\mathrm{d}_3\}$). 
        (b) Irregularly sampled time series is re-represented by aggregating 
        observations $s_m$, which is measured at same time point $\mathrm{t}$. This 
        concept can represent comprehensive status of $S^t$ changing over time $t$. 
        (c) Each variable $d_m$ is embedded into latent space considering measured 
        value $x_m$, which is changing over time. Then, $S^t$ is represented by 
        aggregating observations' embedding at each time point.}
        \label{fig:overview}
    \end{center}
    \vskip -0.2in
\end{figure*}

\subsection{Embedding representation learning}
The concept of TDE is mainly motivated by graph theory, which involves changing edge 
information (i.e., graph topology) over time \cite{kumar2019,beladev2020}. For example, 
each time graph temporal snapshot is represented by concatenating the time index of the 
graph and context node embedding \cite{beladev2020}. In the healthcare domain, there 
are several approaches to represent time-dependent events in embedding space. Choi et al. 
\citeyearpar{choi2016,choi2017} proposed several methods for representing the visit status 
of patients, changing over time by aggregating the embedding of irregularly measured 
clinical events on electronic health record (EHR). Similarly, \citeauthor{lee2020} 
\citeyearpar{lee2020} suggested harmonized representation learning on the EHR, such 
that a patient visit is represented as a latent space by EHR graph construction and 
aggregating the embedding of each clinical event.

\section{Proposed Method}
\label{sec:method}

\subsection{Preliminary}
\textbf{Problem definition}: 
Following the supervised learning concept, each data instance has irregularly sampled 
time series points as inputs and labels as outputs. The dataset is denoted as 
$\mathcal{D}=\{(S_n,y_n) \vert n=1,\dots,N\}$, where $n$ is an index of instance and 
$N$ represents the total number of data instances. Here, $S_n=\{s_m \vert m=1,\dots,M\}$ 
represents a multivariate irregularly sampled time series, where $m$ is an index of 
observation and $M \coloneqq \lvert S_n \rvert$ represents the total number of 
observations of $n$-th instance, which can vary depending on each data instance $S_n$. 
Each observation is denoted by a tuple $s_m=(t_m,x_m,d_m)$, consisting of a measured 
time, $t_m$; a measured value, $x_m$; and a measured variable indicator, 
$d_m \in \{1,\dots,D\}$, where $D$ represents the total number of variables (i.e., 
$D$-dimensional multivariate time series). In practical situations, depending on the 
type of variable, $s_m$ can be a numerical or categorical variable. Further, $s_m$ can 
be a time-independent variable, in that case usually with meaningless $t_m$ or extra 
indicators. Thus, the multivariate irregularly sampled time series for each instance is 
denoted by $S_n=\{(t_1,x_1,d_1),\dots,(t_M,x_M,d_M)\}$, where $t_m$ in different 
observations can be the same value up to the number of $D$, which means that all 
variables are observed at that time point $t_m$ with the assumption that there is no 
duplicated measured value for each variable. In some special cases, if $M=T \times D$, 
where $T \coloneqq \lvert \{ t_m \vert m=1,\dots,M \} \rvert $ represents a unique number 
of $t_m$, it indicates a fully observed multivariate time series. Furthermore, assuming 
that all instances have identical measured time points, (i.e., aligned sampled), $S_n$ 
can be re-represented by pivoting into a structured matrix $\mathbb{R}^{T \times D}$, 
extended to $\mathbb{R}^{N \times T \times D}$ considering all instances. However, the 
practical observations are mostly irregularly measured with unaligned time intervals, 
(i.e., $M \ll T \times D$, and $M_i=\lvert S_i \rvert$ and $M_j=\lvert S_j \rvert$ are 
different). Additionally, some variables are not measured at all, along with each data 
instance situation, which leads to $D'$-dimensional multivariate time series, where 
$D' < D$. $y_n\in\{1,\dots,C\}$ represents its class label, where $C$ represents the 
total number of class labels. Hereafter, we drop the index $n$ for brevity when there 
is no confusion.

To summarize the problem described above, our goal is to classify each class label of 
instance by irregularly sampled time series of which example. 

\subsection{Temporal Dynamic Embedding}
The overview concept of TDE is depicted in Figure~\ref{fig:overview}. Generally, data 
instance $S$ is mainly represented by a schema of multivariate time series (e.g., 
Figure~\ref{fig:overview} (a)). Thus, a data instance at time point $\mathrm{t}$ 
(i.e., $S^\mathrm{t} \coloneqq \{(t_m,x_m,d_m) \vert t_m=\mathrm{t} \, \wedge \, m=1,\dots,M\}$) 
is represented in a fixed $D$-dimensional space, regardless of the absence of each 
element. Instead of the time series schema, Figure~\ref{fig:overview} (b) depicts $S$ 
as a graph network schema changing its edge relationship over time, such that each 
variable $d_m$ becomes a vertex and the measured value $x_m$ becomes an edge under 
index $m$ based on condition $t_m=\mathrm{t}$. Here, $S^\mathrm{t}$ implies the 
comprehensive status of $S$ at time $\mathrm{t}$. To represent $S^\mathrm{t}$ into 
latent space considering the relationship between measurements, each variable $d_m$ 
is also represented as a latent space considering the measured value $x_m$ 
(Figure~\ref{fig:overview} (c)). That is, $S^\mathrm{t}$ can be represented as a 
latent space by aggregating the embedding vector of each variable at time $\mathrm{t}$.

Here, the success of the paradigm shift in graph topology is determined by 
two elements: 
\begin{enumerate}[label=\arabic*)]
    \item Learning the latent information of each variable.
    \item Representing $S^\mathrm{t}$ considering temporal dynamic property.
\end{enumerate}
Now, a detailed description of each element is given as follows.

\subsubsection{Variable Embedding} \label{sec:var_emb}
First, the variable indicator $d_m$ is represented by a binary vector 
$\bm{d}_i = \left\{ 0,1 \right\}^D$ where the $i$-th index element is only one and 
the others zero if $d_m=i$ (i.e., conversion to one-hot-encoding). Then, $\bm{d}_i$ 
is represented in the latent space as follows:. 
\begin{equation} \label{eq:emb}
    \bm{e}_i = ReLU(W_e \bm{d}_i + b_e)
\end{equation}
where $W_e \in \mathbb{R}^{Z \times D}$, $b_e \in \mathbb{R}^{Z}$ are learnable 
parameters and $Z$ represents the embedding space dimension. Consequently, 
Equation~(\ref{eq:emb}) is actually the same as the look-up embedding table, because 
$\bm{d}_i$ is a one-hot-vector that indicates the column index position on $W_e$. 
Although Equation~(\ref{eq:emb}) is represented by a 1-layer neural network with 
ReLU activation, it can be possible with any variants of multi-layer neural networks 
with constraint, only that the last layer is mapped to $Z$-dimensional space and 
with ReLU activation. This is the reason for the constraint keeping the variable 
embedding positive to clarify the property of the observed measurement values when 
aggregating several embedding vectors. In summary, Equation~(\ref{eq:emb}) makes 
each variable indicator to be projected onto a positive latent space.

\subsubsection{Time Embedding}
Each variable embedding $\bm{e}_i$, which is represented in latent space, is based 
on a continuous temporal dynamic characteristic. To represent the time information 
on latent space, we follow the concept of time embedding \cite{shukla2021}, which 
applies the notion of positional encodings \cite{vaswani2017} to continuous time. 
\begin{equation} \label{eq:time_emb}
    \phi(t)[i] = 
        \begin{cases}
        \omega_0 \cdot t + \beta_0                      & \text{if $i=0$}\\
        \sin \left( \omega_i \cdot t + \beta_i \right)  & \text{if $0 < i < Z$}\\
        \end{cases}
\end{equation}
where $\phi(t)\in\mathbb{R}^Z$ and $\omega,\beta \in \mathbb{R}^Z$ are learnable 
parameters. The linear and periodic terms, $\phi(t)[0]$ and $\phi(t)[i \neq 0]$, 
can capture non-periodic and periodic information, respectively, in a time series. 
With this framework for time embedding, the progression of time $\phi(t+\delta)$ 
can be represented as a linear function of $\phi(t)$.

\begin{figure*}[ht]
    \vskip 0.2in
    \begin{center}
        \centerline{\includegraphics[width=0.7\linewidth]{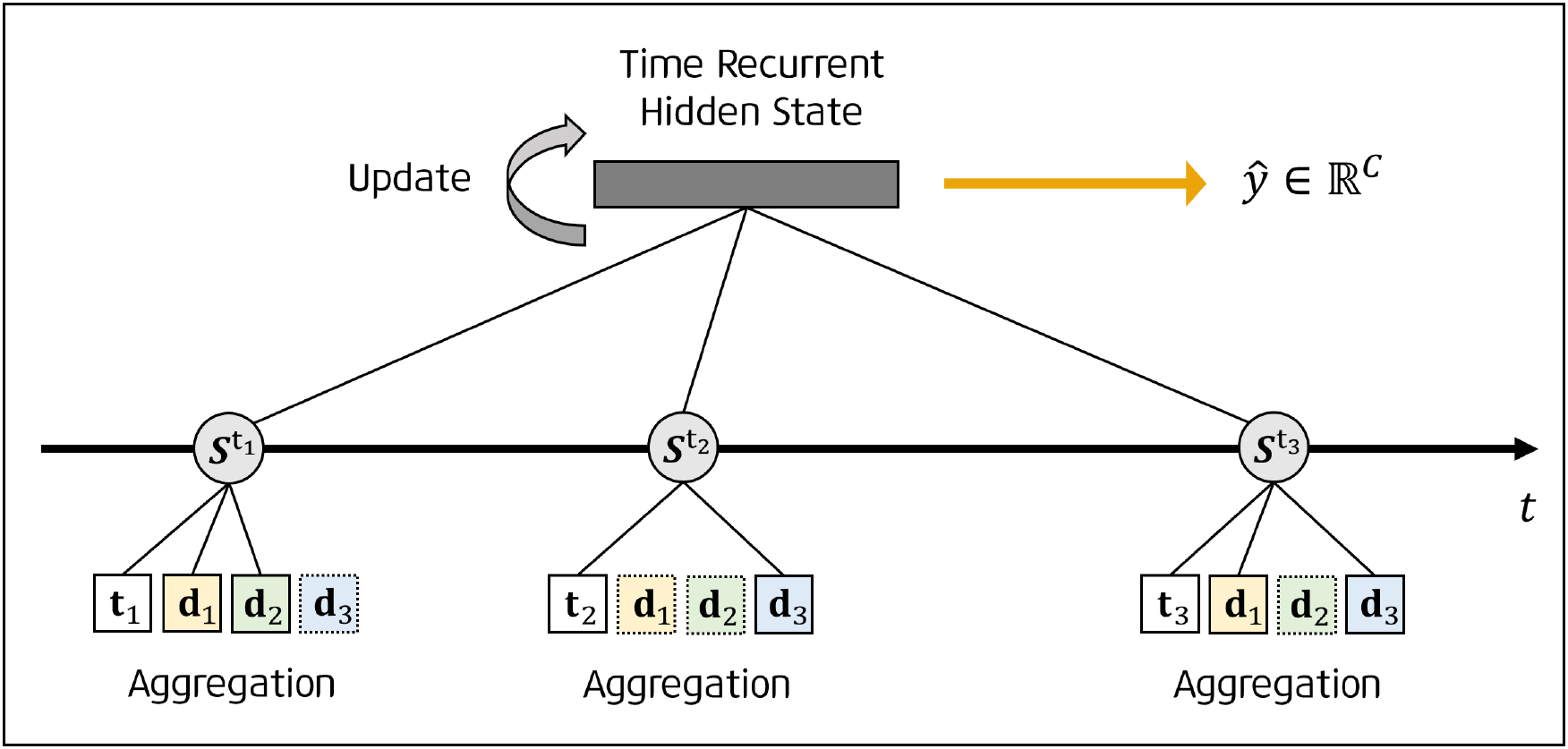}}
        \caption{\textbf{Overall architecture} 
        Each time status is aggregated by each temporal dynamic variable embedding, 
        which leverages update for time-recurrent hidden state. Based on the hidden 
        state of the last observed time step, the probability of a label is calculated.}
        \label{fig:architecture}
    \end{center}
    \vskip -0.2in
\end{figure*}

\subsubsection{Aggregation} \label{sec:agg}
After embedding each variable indicator and time information, the status of the 
data instance at time $t$ can be represented in latent space by aggregating the 
embedding vector of variables measured at that time. The basic form of this equation 
is as follows:
\begin{equation} \label{eq:agg}
    S^t = \sum_{i=1}^{D} {\alpha_i \bm{e}_i + \phi(t)}
\end{equation}
where $\alpha_i$ represents the weight for different variable embeddings considering 
its measured value, and $\phi(t) \in \mathbb{R}^Z$ represents time $t$ information, 
(i.e., time embedding). We established two variants of Equation~(\ref{eq:agg}), 
namely, mean aggregation and attention-based aggregation, along with multi-head 
attention \cite{vaswani2017}. 

\textbf{Mean aggregation}: 
The mean aggregation was designed for the baseline to identify the capability of 
the graph schema. Here, $\alpha_i=x_i/D_t$, where $D_t$ is the number of measured 
variables at time $t$. 

\textbf{Attention-based aggregation}: 
The attention-based aggregation is designed to consider the variable measurement 
value and correlation between the observed variables. 
\begin{equation} \label{eq:agg_attn}
    S^t = f\left( \textrm{Concat}( \sum_{i=1}^{D} {\alpha_{ih} V_{ih}} )_{h=1}^{H} \right) + \phi(t)
\end{equation}
where $H$ is the number of multi-heads, and $f(\cdot)$ is a multi-layer neural 
network. Here, $\alpha_i V_i$, which drops the index $h$ for brevity, is as follows: 
\begin{equation} \label{eq:attn}
\begin{split}
    \alpha_i V_i & = \sum_{j=1}^{D} {attn(\bm{e}_i, \bm{e}_j, x_j) V_i} \\
                 & = \sum_{j=1}^{D} {x_j \frac{Q_i^T K_j}{\sqrt{D_k}} V_i} \\
\end{split}
\end{equation}
where $attn(\cdot)$ is the attention module, which takes three arguments as query, 
key, and value, that calculates the importance of each observed variable. Here, 
$Q_i = W_q \bm{e}_i$, $K_j = W_k \bm{e}_j$, $V_i = W_v \bm{e}_i$, and 
$W_q \in \mathbb{R}^{D_q \times Z}$, $W_k \in \mathbb{R}^{D_k \times Z}$, 
$W_v \in \mathbb{R}^{D_v \times Z}$, and $D_q,D_k,D_v$ represent the dimensions 
of the query, key, and value, respectively, and $D_q=D_k$. As shown in 
Equations~(\ref{eq:agg_attn}) and (\ref{eq:attn}), the basic framework for 
attention-based aggregation follows multi-head attention. However, there are 
three significant differences between our proposed method and original 
multi-head attention method. (1) The softmax wrapper on the attention score was 
discarded. Because each time step has a different number of measurements, which 
leads to different numbers of embedding vectors $V_i$ for aggregation, $S^t$ 
cannot be expressed by a fairly normalized weight on each embedding vector with 
a softmax output. For example, we suppose the case of only a single measured 
variable and the case of 100. With the softmax layer, the attention score of the 
former case is always 1.0, while in the latter case, the impact of each variable 
is diminished because there are too many candidates to be allocated a weight-score 
limited sum up to 1.0. Through the ablation study in Appendix~\ref{sec:app_ablation}, 
we identified that the non-softmax attention score performed better on our task; 
therefore we discarded the softmax layer. (2) The attention score on $\bm{e}_i$ is 
determined by considering $\bm{e}_j$ and $x_j$ and the sum of its results over $j$. 
It is possible to consider the correlation effect between each variable when 
calculating the attention score. Further, it can capture temporal dynamic information 
through $x_j$ term, which is the measured value that changes over time. (3) $S^t$ is 
calculated at every time step, based on the attention mechanism between each measured 
variable. Thus, this architecture can capture temporal dynamic characteristics for 
irregularly sampled time series in real time. 

\subsection{Learning framework}
\subsubsection{Learning Temporal Status} \label{sec:temp_status}
We represented the temporal status based on the aggregation of variable embedding. 
Now, we leverage temporal dynamic status into a time recurrent hidden state, which 
can be considered as local status and global status, respectively. The time 
recurrent hidden state (i.e., global status) was updated by the temporal dynamic 
status (i.e., local status), which contains information only measured at a particular 
time. Generally, a hidden state is developed along with the variants of the RNN model. 
In this study, we developed a time-recurrent hidden state as follows: 
\begin{equation} \label{eq:gru}
\begin{gathered}
    r_t = \sigma(W_r S^t + U_r h_{t-1} + b_r) \\
    z_t = \sigma(W_z S^t + U_z h_{t-1} + b_z) \\
    \tilde{h}_t = \tanh \left( W_h S^t + U_h (r_t \odot h_{t-1}) + b \right) \\
    h_t = z_t \odot \tilde{h}_t + (1-z_t) \odot h_{t-1} \\
\end{gathered}
\end{equation}
where $\odot$ represents element-wise multiplication. Equation~(\ref{eq:gru}) is a 
well-known GRU model \cite{cho2014}. Thus, the hidden state can be determined by 
$S^t$ whether updated or not, such that it contains significant information. 

\subsubsection{Loss Function}
The overall architecture is depicted in 
Figure~\ref{fig:architecture}. Along with the supervised learning concept, we 
trained all learnable parameters in architecture to maximize the probability 
of correctly predicting $y_n$, given the hidden state of the last observed time 
step, which is described above. The loss function is defined as: 
\begin{equation} \label{eq:loss}
    \mathcal{L}(\theta,\psi) = \underset{\theta,\psi}{\arg \min} 
                               - \frac{1}{N} \sum_{n=1}^{N} {\log p_\theta (y_n \vert h_\psi^T(S_n))}
\end{equation}
where $h_\psi^T(S_n)$ is the hidden state of the last time step, which consists of 
all learnable parameters related to the hidden state, $\psi$, and inputs, $S_n$. 
$p_\theta$ is a neural network classifier with learnable parameters $\theta$ for 
predicting the probability of $y_n$ given $h_\psi^T(S_n)$. We used $p_\theta$ as a 
simple, fully connected neural network for classification.

\section{Experiments}
\label{sec:experiments}

In this section, we describe all our experiments and implementations. Although our 
proposed method, TDE, is applicable to a wide range of applications following 
irregularly sampled time series, we focused our experiments on benchmark clinical 
tasks. The implementation is publicly available\footnote{Implementation available at: 
\url{https://github.com/mnchl-kim/TDE-ICML2022}}.

\subsection{Datasets} \label{sec:datasets}
\textbf{The PhysioNet Challenge 2012}: 
\emph{The PhysioNet Computing in Cardiology Challenge 2012} \cite{silva2012} dataset 
is a publicly available record of 12,000 intensinve care unit (ICU) patients, obtained 
within the first 48 hours after the first ICU admission. The aim of this challenge is 
the patient-specific prediction of in-hospital mortality. The in-hospital mortality 
ratio was 14\%, which was imbalanced. With a conventional structured representation 
of a time series, its sparsity was approximately 95\% missing ratio. By discretizing 
into a fixed hourly bin, which can cause loss of observed information, its sparsity 
was slightly reduced, but it was still sufficiently severe, i.e., approximately 81\% 
missing ratio. Considering the original challenge dataset characteristics, we used 
set A, B, and C as the training, validation, and test datasets, respectively, which 
consisted data of 4,000 patients.

\textbf{MIMIC-\Romannum{3}}:
\emph{Medical Information Mart for Intensive Care - \Romannum{3}} \cite{johnson2016} 
 dataset is a publicly available de-identified EHR under credentialed certification. 
 MIMIC-\Romannum{3} recorded data of 58,976 patients admitted into the Beth Israel 
 Deaconess Medical Center between 2001 and 2012. Following the task of 
 \emph{The PhysioNet Challenge 2012}, we designed a prediction of in-hospital mortality 
 task with measurements within the first 48 hours after the first ICU admission. After 
 cohort selection and data preprocessing, we took 26,966 hospital admissions, of which 
 the in-hospital mortality ratio was 14\%. With a conventional structured representation 
 of a time series, its sparsity was approximately 90\% missing ratio. By discretizing 
 into a fixed hourly bin, the sparsity was approximately 51\% missing ratio.

\textbf{The PhysioNet Challenge 2019}: 
\emph{The PhysioNet Computing in Cardiology Challenge 2019} \cite{reyna2019} dataset 
involved ICU patients in three different hospital systems. The aim of this challenge 
was the early detection of sepsis using clinical data. With a publicly available 
record of 40,336 patients from set A and B, the task is the hourly prediction of 
sepsis onset 6 hours before the clinical prediction of sepsis, which is defined by the 
Sepsis-3 guidelines \cite{singer2016}. The sepsis onset ratio was 7\%, but considering 
all online prediction as separate cases, the sepsis onset ratio was only 2\%. With a 
conventional structured representation of a time series, its sparsity was approximately 
80\% missing ratio. In this challenge, data were already provided by discretizing into 
fixed hourly bins.

All detailed procedures of data preprocessing, the variable list, and experiment settings of the three 
datasets are described in the Appendix~\ref{sec:app_datasets}.

\begin{table*}[ht]
\caption{Performance comparison for classification tasks}
\label{tab:results}
    \begin{center}
    \begin{adjustbox}{width=\linewidth}
    \begin{small}
        \begin{threeparttable}
        \begin{tabular}{lccccccr}
            \toprule
            \multirow{2}[2]{*}{Model} & \multicolumn{2}{c}{PhysioNet 2012} & \multicolumn{2}{c}{MIMIC-\Romannum{3}} & \multicolumn{2}{c}{PhysioNet 2019} & \multirow{2}[2]{*}{Time\tnote{*}} \\
            \cmidrule(lr){2-3} \cmidrule(lr){4-5} \cmidrule(lr){6-7}
            {} & AUROC & AUPRC & AUROC & AUPRC & AUROC & AUPRC & {} \\
            \midrule
            RNN         & {0.749 $\pm$ 0.011} & {0.370 $\pm$ 0.020} & {0.560 $\pm$ 0.012} & {0.168 $\pm$ 0.008} & {0.729 $\pm$ 0.009} & {0.045 $\pm$ 0.003} & {2.1} \\
            LSTM        & {0.816 $\pm$ 0.009} & {0.440 $\pm$ 0.021} & {0.580 $\pm$ 0.010} & {0.176 $\pm$ 0.008} & {0.784 $\pm$ 0.009} & {0.077 $\pm$ 0.005} & {1.5} \\
            GRU         & {0.828 $\pm$ 0.009} & {0.466 $\pm$ 0.022} & {0.784 $\pm$ 0.009} & {0.392 $\pm$ 0.019} & {0.795 $\pm$ 0.008} & {0.077 $\pm$ 0.005} & {4.3} \\
            GRU-D       & {0.847 $\pm$ 0.008} & {0.505 $\pm$ 0.021} & {0.789 $\pm$ 0.009} & {0.423 $\pm$ 0.020} & {\textbf{0.823 $\pm$ 0.007}} & {\textbf{0.095 $\pm$ 0.006}} & {11.7} \\
            BRITS       & {0.834 $\pm$ 0.008} & {0.466 $\pm$ 0.022} & {0.811 $\pm$ 0.008} & {0.422 $\pm$ 0.019} & {-} & {-} & {17.1} \\
            mTAND-Enc   & {0.852 $\pm$ 0.008} & {0.516 $\pm$ 0.021} & {0.767 $\pm$ 0.010} & {0.409 $\pm$ 0.019} & {-} & {-} & {9.6} \\
            mTAND-Full  & {\textbf{0.854 $\pm$ 0.001}} & {0.522 $\pm$ 0.001} & {0.771 $\pm$ 0.002} & {0.409 $\pm$ 0.003} & {-} & {-} & {-} \\
            \midrule
            TDE-Mean    & {0.842 $\pm$ 0.008} & {0.524 $\pm$ 0.022} & {0.812 $\pm$ 0.008} & {0.458 $\pm$ 0.020} & {0.819 $\pm$ 0.007} & {0.082 $\pm$ 0.006} & {5.3} \\
            TDE-Attn    & {0.848 $\pm$ 0.009} & {\textbf{0.532 $\pm$ 0.022}} & {\textbf{0.818 $\pm$ 0.008}} & {\textbf{0.475 $\pm$ 0.019}} & {0.819 $\pm$ 0.007} & {\textbf{0.095 $\pm$ 0.008}} & {6.4} \\
            \bottomrule
        \end{tabular}
        \begin{tablenotes}\footnotesize
        \item * Time per epoch ($s/epoch$) : measured by with PhysioNet 2012 dataset.
        \end{tablenotes}
        \end{threeparttable}
    \end{small}
    \end{adjustbox}
    \end{center}
\end{table*}

\subsection{Models}
The proposed method was suggested by the constraint of data sparsity with a 
conventional structured representation. Based on variable embedding, we 
suggested two variants of the aggregation method for TDE: TDE-Mean and TDE-Attn. 
To identify the capability of the proposed method, we compared several 
imputation-based methods with conventional structured representation and recent 
end-to-end methods. All baseline methods compared with the proposed method are 
described below. 
\begin{itemize}
    \item \textbf{RNN} : RNN with a conventional structured representation, which 
    replaced missing as zero imputation. Because of input normalization, zero 
    imputation was equal to the mean imputation. 
    \item \textbf{LSTM} : LSTM with conventional structured representation, which 
    replaced missing as zero imputation \cite{hochreiter1997}. 
    \item \textbf{GRU} : GRU with conventional structured representation, which 
    replaced missing as zero imputation \cite{cho2014}. 
    \item \textbf{GRU-D} : Combining hidden state decay and input decay, which is 
    developed by missing imputation with a weighted average of the last observed 
    measurement and global mean \cite{che2018}. 
    \item \textbf{BRITS} : Missing imputation by a bidirectional recurrent unit, 
    which is a generalized version of GRU-D \cite{cao2018}.
    \item \textbf{mTAND-Enc} : End-to-end method based on interpolation at several 
    reference points using discretized multi-time attention network (mTAND) module 
    \cite{shukla2021}.
    \item \textbf{mTAND-Full} : mTAND model with a VAE framework \cite{shukla2021}.
    \item \textbf{TDE-Mean} : Mean aggregation for TDE (Equation~\ref{eq:agg}).
    \item \textbf{TDE-Attn} : Attention-based aggregation for TDE 
    (Equation~\ref{eq:agg_attn}).
\end{itemize}

\subsection{Results}
Table~\ref{tab:results} compares the classification performance on \emph{The PhysioNet 
Challenge 2012}, \emph{MIMIC-\Romannum{3}}, and \emph{The PhysioNet Challenge 2019}, 
and training runtime performance on \emph{The PhysioNet Challenge 2012}. Because of 
imbalanced datasets, we selected the performance metrics as area under the receiver 
operating characteristics curve (AUROC) and area under the precision-recall curve 
(AUPRC), which was able to capture the classifying ability for rare class.

\textbf{The PhysioNet Challenge 2012}: 
The TDE-based model achieved the best performance on AUPRC, although the mTAND-based 
model achieved slightly better performance than the TDE-based model on AUROC. In the 
case of a serious imbalance problem, the AUPRC metric is more important because it can 
capture the classifying ability whether the model can predict correctly for rare cases 
of in-hospital death. 

Moreover, the training runtime per epoch in seconds was measured by all methods under 
the same environments, i.e., NVIDIA Geforce RTX 2080 Ti, with a fixed batch size. Under 
identical circumstances, mTAND-Full model required memory that exceeded the capacity 
of the GPU memory; thus, it could not be measured. Compared to recent state-of-the-art 
methods, TDE was two or three times faster, and TDE was rather closer to GRU, which is 
also used as a component of TDE. This means that the variable embedding and aggregation 
procedure had little effect on runtime but significantly improved classification 
performance.

\textbf{MIMIC-\Romannum{3}}: 
This result shows that TDE had noticeable improvement in AUROC and AUPRC compared 
to the other methods. RNN and LSTM failed to learn meaningful classifying capability, 
whereas GRU achieved similar performance to GRU-D, which means that decay-based 
imputation had little effect on classifying class labels in this dataset. Unlike 
the PhysioNet 2012 results, mTAND could not achieve outstanding performance, and 
BRITS, which is not as good for PhysioNet 2012, achieved the second-best performance. 
There was weak consistency in recent state-of-the-art methods based on different 
datasets.

\textbf{The PhysioNet Challenge 2019}: 
Because this task requires online prediction, imputation-based methods are not 
precisely suitable. BRITS and mTAND should precede bidirectional imputation or 
interpolation procedures with already measured observations; thus, all observations, 
even after the online prediction point, can cause leakage of future information. 
\citeauthor{horn2020} \citeyearpar{horn2020} proved the risk of imputation-based 
methods for online prediction. To avoid this risk, each instance can be reprocessed 
for future information to be unavailable. However, this process is a secondary, 
additional burden for online prediction. In this study, we considered only other 
baseline methods that have no constraints for online prediction, and TDE and GRU-D 
outperformed AUROC and AUPRC. GRU-D was slightly better than TDE, but it could be 
explained by the property of data, which was already given by discretizing into a 
fixed regular hourly bin. Therefore, the problem caused by irregularly sampled 
time series was alleviated, which reduced the effect of temporal dynamic embedding 
aggregation. Furthermore, because of the serious imbalance problem, the AUPRC 
for all methods was less than 0.1.

\begin{figure*}[ht]
    \vskip 0.2in
    \begin{center}
        \centerline{\includegraphics[width=0.8\linewidth]{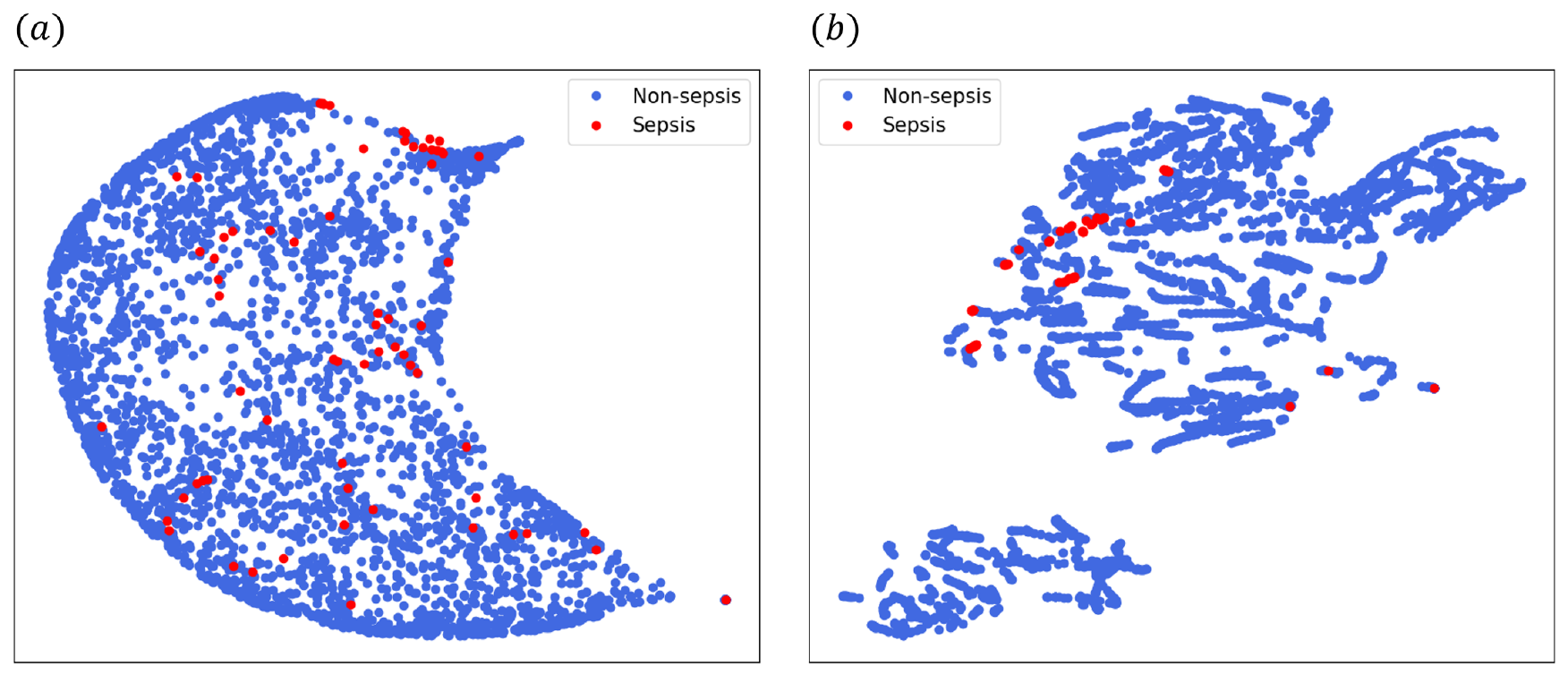}}
        \caption{\textbf{Two-dimensional embedding representation using t-SNE for 
        \emph{The PhysioNet Challenge 2019}.} (a) Representation of the aggregation 
        between the observed variable embedding, measurements and time embedding at each 
        time. (b) Representation of the recurrent hidden state of the last time step.}
        \label{fig:embedding}
    \end{center}
    \vskip -0.2in
\end{figure*}

\subsection{Embedding representation}
To verify the effect of TDE, Figure~\ref{fig:embedding} 
illustrates a two-dimensional embedding representation using t-distributed stochastic 
neighbor embedding (t-SNE) \cite{maaten2008} learned from \emph{The PhysioNet Challenge 
2019}. The TDE model learned implicit relationships between all variables sets without 
any explicit supervision of the variable relationships. Figure~\ref{fig:embedding} 
(a) shows the aggregation of the observed variable embedding, their measurement values 
and time embedding at each time point (Section~\ref{sec:agg}). Each time step aggregation 
only considers current observations without previous time step information, which can 
be regarded as local status. The blue dots indicate non-sepsis patients, while the red 
dots indicate sepsis patients. Figure~\ref{fig:embedding} (b) shows the embedding of 
the recurrent hidden state of the last time step, considering the information of the 
previous time step, which can be regarded as global status (Section~\ref{sec:temp_status}). 
Although there was a weak distinct character to distinguish class labels with only 
local status information, we could identify a sepsis-onset cluster with global status, 
which considers previous time step information. Because most patients admitted to the 
ICU are evaluated the risk state through the overall condition after hospital admission 
not fragmental information of the current state, this results implied that accumulation 
of information over time can disentangle representation of the risk state.

\section{Discussion and Conclusion}
\label{sec:disc_conc}

In this study, we proposed a novel approach to temporal dynamic embedding for 
irregularly sampled time series, which is a common yet critical problem in many 
practical applications. Our method considers only observed data at each time point 
regardless of the number of subsets of variables measured, which is a critical 
problem for conventional fixed structured representation. This approach enables 
neural network based models to receive data that change the number of variables 
over time. Our results on three clinical tasks, including PhysioNet 2012, 
MIMIC-\Romannum{3}, and PhysioNet 2019, show that the proposed TDE model performed 
competitively or better than the imputation-based baseline such as RNN, LSTM, and 
GRU with conventional fixed structured representation, and several recent 
state-of-the-art methods such as GRU-D, BRITS, and mTAND. With competitively or 
significantly improved classification performance, the runtime for training optimal 
model parameters was also reduced by more than two or three times, compared to recent 
methods. We showed that variable embedding and aggregation procedure has little effect 
on runtime, but significantly improves classification performance compared to runtime 
for GRU, which is also used as a component of TDE.

Along with the paradigm shift of time-series into graph topology, we focused on two 
elements: learning the latent information of each variable and representing each time 
status considering temporal dynamic property. In the latter case, we suggest two 
aggregation methods for representing each time status using mean aggregation and 
attention-based aggregation of the observed variable embedding,  its measurement 
value, and time embedding. In the former case, although we did not design an explicit 
way to learn the relationship and similarity of a variable, e.g., Word2Vec 
\cite{mikolov2013}, or Med2Vec \cite{choi2016}, it is sufficient to show competitively 
or improved classification performance with a naive trainable lookup table for 
variable embedding. In future research, we plan to design self-supervised learning 
for the relationships between entire variable sets in time series. It is expected 
that better performance on classification tasks can be accomplished and the semantic 
relationships between several time series signals can be determined, contributing to 
self-supervised learning on time series.

\clearpage
\bibliographystyle{main}
\bibliography{main}

\begin{thebibliography}{43}
\providecommand{\natexlab}[1]{#1}
\providecommand{\url}[1]{\texttt{#1}}
\expandafter\ifx\csname urlstyle\endcsname\relax
  \providecommand{\doi}[1]{doi: #1}\else
  \providecommand{\doi}{doi: \begingroup \urlstyle{rm}\Url}\fi

\bibitem[Ansley \& Kohn(1984)Ansley and Kohn]{ansley1984}
Ansley, C.~F. and Kohn, R.
\newblock {On the estimation of ARIMA Models with Missing Values}.
\newblock In \emph{Time Series Analysis of Irregularly Observed Data}, pp.\  9--37. Springer New York, 1984.

\bibitem[Beladev et~al.(2020)Beladev, Rokach, Katz, Guy, and Radinsky]{beladev2020}
Beladev, M., Rokach, L., Katz, G., Guy, I., and Radinsky, K.
\newblock {TdGraphEmbed: Temporal Dynamic Graph-Level Embedding}.
\newblock In \emph{Proceedings of the 29th ACM International Conference on Information and Knowledge Management}, pp.\  55–64. Association for Computing Machinery, 2020.

\bibitem[Cao et~al.(2018)Cao, Wang, Li, Zhou, Li, and Li]{cao2018}
Cao, W., Wang, D., Li, J., Zhou, H., Li, L., and Li, Y.
\newblock {BRITS: Bidirectional Recurrent Imputation for Time Series}.
\newblock In \emph{Advances in Neural Information Processing Systems}, volume~31, pp.\  6776--6786, 2018.

\bibitem[Che et~al.(2018)Che, Purushotham, Cho, Sontag, and Liu]{che2018}
Che, Z., Purushotham, S., Cho, K., Sontag, D., and Liu, Y.
\newblock {Recurrent Neural Networks for Multivariate Time Series with Missing Values}.
\newblock \emph{Scientific Reports}, 8\penalty0 (1):\penalty0 1--12, 2018.

\bibitem[Cho et~al.(2014)Cho, van Merri{\"e}nboer, Gulcehre, Bahdanau, Bougares, Schwenk, and Bengio]{cho2014}
Cho, K., van Merri{\"e}nboer, B., Gulcehre, C., Bahdanau, D., Bougares, F., Schwenk, H., and Bengio, Y.
\newblock {Learning Phrase Representations using RNN Encoder--Decoder for Statistical Machine Translation}.
\newblock In \emph{Proceedings of the 2014 Conference on Empirical Methods in Natural Language Processing}, pp.\  1724--1734. Association for Computational Linguistics, 2014.

\bibitem[Choi et~al.(2016)Choi, Bahadori, Searles, Coffey, Thompson, Bost, Tejedor-Sojo, and Sun]{choi2016}
Choi, E., Bahadori, M.~T., Searles, E., Coffey, C., Thompson, M., Bost, J., Tejedor-Sojo, J., and Sun, J.
\newblock {Multi-Layer Representation Learning for Medical Concepts}.
\newblock In \emph{Proceedings of the 22nd ACM SIGKDD International Conference on Knowledge Discovery and Data Mining}, pp.\  1495–1504. Association for Computing Machinery, 2016.

\bibitem[Choi et~al.(2017)Choi, Bahadori, Song, Stewart, and Sun]{choi2017}
Choi, E., Bahadori, M.~T., Song, L., Stewart, W.~F., and Sun, J.
\newblock {GRAM: Graph-Based Attention Model for Healthcare Representation Learning}.
\newblock In \emph{Proceedings of the 23rd ACM SIGKDD International Conference on Knowledge Discovery and Data Mining}, pp.\  787–795. Association for Computing Machinery, 2017.

\bibitem[Fawaz et~al.(2019)Fawaz, Forestier, Weber, Idoumghar, and Muller]{fawaz2019}
Fawaz, H.~I., Forestier, G., Weber, J., Idoumghar, L., and Muller, P.-A.
\newblock {Deep learning for time series classification: a review}.
\newblock \emph{Data mining and knowledge discovery}, 33\penalty0 (4):\penalty0 917--963, 2019.

\bibitem[Friedman et~al.(2001)Friedman, Hastie, Tibshirani, et~al.]{friedman2001}
Friedman, J., Hastie, T., Tibshirani, R., et~al.
\newblock \emph{{The elements of statistical learning}}, volume~1.
\newblock Springer series in statistics New York, 2001.

\bibitem[Garc\'{\i}a-Laencina et~al.(2010)Garc\'{\i}a-Laencina, Sancho-G\'{o}mez, and Figueiras-Vidal]{garcia-laencina2010}
Garc\'{\i}a-Laencina, P.~J., Sancho-G\'{o}mez, J.-L., and Figueiras-Vidal, A.~R.
\newblock {Pattern Classification with Missing Data: A Review}.
\newblock \emph{Neural Comput. Appl.}, 19\penalty0 (2):\penalty0 263–282, 2010.

\bibitem[Goodfellow et~al.(2014)Goodfellow, Pouget-Abadie, Mirza, Xu, Warde-Farley, Ozair, Courville, and Bengio]{goodfellow2014}
Goodfellow, I., Pouget-Abadie, J., Mirza, M., Xu, B., Warde-Farley, D., Ozair, S., Courville, A., and Bengio, Y.
\newblock {Generative Adversarial Nets}.
\newblock In \emph{Advances in Neural Information Processing Systems}, volume~27, pp.\  2672--2680, 2014.

\bibitem[Hochreiter \& Schmidhuber(1997)Hochreiter and Schmidhuber]{hochreiter1997}
Hochreiter, S. and Schmidhuber, J.
\newblock {Long Short-Term Memory}.
\newblock \emph{Neural Computation}, 9\penalty0 (8):\penalty0 1735--1780, 11 1997.

\bibitem[Horn et~al.(2020)Horn, Moor, Bock, Rieck, and Borgwardt]{horn2020}
Horn, M., Moor, M., Bock, C., Rieck, B., and Borgwardt, K.~M.
\newblock {Set Functions for Time Series}.
\newblock In \emph{Proceedings of the 37th International Conference on Machine Learning}, volume 119 of \emph{Proceedings of Machine Learning Research}, pp.\  4353--4363. {PMLR}, 2020.

\bibitem[James et~al.(2013)James, Witten, Hastie, and Tibshirani]{james2013}
James, G., Witten, D., Hastie, T., and Tibshirani, R.
\newblock \emph{{An introduction to statistical learning}}, volume 112.
\newblock Springer, 2013.

\bibitem[Johnson et~al.(2016)Johnson, Pollard, Shen, Li-Wei, Feng, Ghassemi, Moody, Szolovits, Celi, and Mark]{johnson2016}
Johnson, A.~E., Pollard, T.~J., Shen, L., Li-Wei, H.~L., Feng, M., Ghassemi, M., Moody, B., Szolovits, P., Celi, L.~A., and Mark, R.~G.
\newblock {MIMIC-III, a freely accessible critical care database}.
\newblock \emph{Scientific data}, 3\penalty0 (1):\penalty0 1--9, 2016.

\bibitem[Kingma \& Welling(2014)Kingma and Welling]{kingma2014}
Kingma, D.~P. and Welling, M.
\newblock {Auto-Encoding Variational Bayes}.
\newblock In \emph{2nd International Conference on Learning Representations}, 2014.

\bibitem[Koren et~al.(2009)Koren, Bell, and Volinsky]{koren2009}
Koren, Y., Bell, R., and Volinsky, C.
\newblock {Matrix factorization techniques for recommender systems}.
\newblock \emph{Computer}, 42\penalty0 (8):\penalty0 30--37, 2009.

\bibitem[Kumar et~al.(2019)Kumar, Zhang, and Leskovec]{kumar2019}
Kumar, S., Zhang, X., and Leskovec, J.
\newblock {Predicting Dynamic Embedding Trajectory in Temporal Interaction Networks}.
\newblock In \emph{Proceedings of the 25th ACM SIGKDD International Conference on Knowledge Discovery and Data Mining}, pp.\  1269–1278. Association for Computing Machinery, 2019.

\bibitem[Lee et~al.(2020)Lee, Jiang, and Yu]{lee2020}
Lee, D., Jiang, X., and Yu, H.
\newblock {Harmonized representation learning on dynamic EHR graphs}.
\newblock \emph{Journal of Biomedical Informatics}, 106:\penalty0 103426, 2020.

\bibitem[Li et~al.(2019)Li, Jiang, and Marlin]{li2019}
Li, S. C.-X., Jiang, B., and Marlin, B.
\newblock {Learning from Incomplete Data with Generative Adversarial Networks}.
\newblock In \emph{International Conference on Learning Representations}, 2019.

\bibitem[Lipton et~al.(2016)Lipton, Kale, and Wetzel]{lipton2016}
Lipton, Z.~C., Kale, D., and Wetzel, R.
\newblock {Directly Modeling Missing Data in Sequences with RNNs: Improved Classification of Clinical Time Series}.
\newblock In \emph{Proceedings of the 1st Machine Learning for Healthcare Conference}, volume~56 of \emph{Proceedings of Machine Learning Research}, pp.\  253--270. PMLR, 2016.

\bibitem[Luo et~al.(2019)Luo, Zhang, Cai, and Yuan]{luo2019}
Luo, Y., Zhang, Y., Cai, X., and Yuan, X.
\newblock {E²GAN: End-to-End Generative Adversarial Network for Multivariate Time Series Imputation}.
\newblock In \emph{Proceedings of the Twenty-Eighth International Joint Conference on Artificial Intelligence}, pp.\  3094--3100. International Joint Conferences on Artificial Intelligence Organization, 2019.

\bibitem[Marlin et~al.(2012)Marlin, Kale, Khemani, and Wetzel]{marlin2012}
Marlin, B.~M., Kale, D.~C., Khemani, R.~G., and Wetzel, R.~C.
\newblock {Unsupervised Pattern Discovery in Electronic Health Care Data Using Probabilistic Clustering Models}.
\newblock In \emph{Proceedings of the 2nd ACM SIGHIT International Health Informatics Symposium}, pp.\  389--398. Association for Computing Machinery, 2012.

\bibitem[Mikolov et~al.(2013)Mikolov, Sutskever, Chen, Corrado, and Dean]{mikolov2013}
Mikolov, T., Sutskever, I., Chen, K., Corrado, G.~S., and Dean, J.
\newblock Distributed representations of words and phrases and their compositionality.
\newblock In \emph{Advances in Neural Information Processing Systems}, volume~26, 2013.

\bibitem[Purushotham et~al.(2018)Purushotham, Meng, Che, and Liu]{purushotham2018}
Purushotham, S., Meng, C., Che, Z., and Liu, Y.
\newblock {Benchmarking deep learning models on large healthcare datasets}.
\newblock \emph{Journal of Biomedical Informatics}, 83:\penalty0 112--134, 2018.

\bibitem[Ravuri et~al.(2021)Ravuri, Lenc, Willson, Kangin, Lam, Mirowski, Fitzsimons, Athanassiadou, Kashem, Madge, Prudden, Mandhane, Clark, Brock, Simonyan, Hadsell, Robinson, Clancy, Arribas, and Mohamed]{ravuri2021}
Ravuri, S., Lenc, K., Willson, M., Kangin, D., Lam, R., Mirowski, P., Fitzsimons, M., Athanassiadou, M., Kashem, S., Madge, S., Prudden, R., Mandhane, A., Clark, A., Brock, A., Simonyan, K., Hadsell, R., Robinson, N., Clancy, E., Arribas, A., and Mohamed, S.
\newblock {Skilful precipitation nowcasting using deep generative models of radar}.
\newblock \emph{Nature}, 597\penalty0 (7878):\penalty0 672--677, Sep 2021.

\bibitem[Reyna et~al.(2019)Reyna, Josef, Seyedi, Jeter, Shashikumar, Brandon~Westover, Sharma, Nemati, and Clifford]{reyna2019}
Reyna, M.~A., Josef, C., Seyedi, S., Jeter, R., Shashikumar, S.~P., Brandon~Westover, M., Sharma, A., Nemati, S., and Clifford, G.~D.
\newblock {Early Prediction of Sepsis from Clinical Data: the PhysioNet/Computing in Cardiology Challenge 2019}.
\newblock In \emph{2019 Computing in Cardiology}, pp.\  1--4, 2019.

\bibitem[Sezer et~al.(2020)Sezer, Gudelek, and Ozbayoglu]{sezer2020}
Sezer, O.~B., Gudelek, M.~U., and Ozbayoglu, A.~M.
\newblock {Financial time series forecasting with deep learning : A systematic literature review: 2005–2019}.
\newblock \emph{Applied Soft Computing}, 90:\penalty0 106181, 2020.

\bibitem[Shukla \& Marlin(2019)Shukla and Marlin]{shukla2019}
Shukla, S.~N. and Marlin, B.
\newblock {Interpolation-Prediction Networks for Irregularly Sampled Time Series}.
\newblock In \emph{International Conference on Learning Representations}, 2019.

\bibitem[Shukla \& Marlin(2021{\natexlab{a}})Shukla and Marlin]{shukla2021}
Shukla, S.~N. and Marlin, B.
\newblock {Multi-Time Attention Networks for Irregularly Sampled Time Series}.
\newblock In \emph{International Conference on Learning Representations}, 2021{\natexlab{a}}.

\bibitem[Shukla \& Marlin(2021{\natexlab{b}})Shukla and Marlin]{shukla2021survey}
Shukla, S.~N. and Marlin, B.~M.
\newblock {A Survey on Principles, Models and Methods for Learning from Irregularly Sampled Time Series}.
\newblock \emph{arXiv preprint arXiv:2012.00168v2}, 2021{\natexlab{b}}.

\bibitem[Silva et~al.(2012)Silva, Moody, Scott, Celi, and Mark]{silva2012}
Silva, I., Moody, G., Scott, D.~J., Celi, L.~A., and Mark, R.~G.
\newblock {Predicting in-hospital mortality of ICU patients: The PhysioNet/Computing in cardiology challenge 2012}.
\newblock In \emph{2012 Computing in Cardiology}, pp.\  245--248, 2012.

\bibitem[Singer et~al.(2016)Singer, Deutschman, Seymour, Shankar-Hari, Annane, Bauer, Bellomo, Bernard, Chiche, Coopersmith, Hotchkiss, Levy, Marshall, Martin, Opal, Rubenfeld, van~der Poll, Vincent, and Angus]{singer2016}
Singer, M., Deutschman, C.~S., Seymour, C.~W., Shankar-Hari, M., Annane, D., Bauer, M., Bellomo, R., Bernard, G.~R., Chiche, J.-D., Coopersmith, C.~M., Hotchkiss, R.~S., Levy, M.~M., Marshall, J.~C., Martin, G.~S., Opal, S.~M., Rubenfeld, G.~D., van~der Poll, T., Vincent, J.-L., and Angus, D.~C.
\newblock {The Third International Consensus Definitions for Sepsis and Septic Shock (Sepsis-3)}.
\newblock \emph{JAMA}, 315\penalty0 (8):\penalty0 801--810, 02 2016.

\bibitem[Srivastava et~al.(2014)Srivastava, Hinton, Krizhevsky, Sutskever, and Salakhutdinov]{srivastava2014}
Srivastava, N., Hinton, G., Krizhevsky, A., Sutskever, I., and Salakhutdinov, R.
\newblock {Dropout: A Simple Way to Prevent Neural Networks from Overfitting}.
\newblock \emph{Journal of Machine Learning Research}, 15\penalty0 (56):\penalty0 1929--1958, 2014.

\bibitem[Suo et~al.(2020)Suo, Zhong, Xun, Sun, Chen, and Zhang]{suo2020}
Suo, Q., Zhong, W., Xun, G., Sun, J., Chen, C., and Zhang, A.
\newblock {GLIMA: Global and Local Time Series Imputation with Multi-directional Attention Learning}.
\newblock In \emph{2020 IEEE International Conference on Big Data}, pp.\  798--807, 2020.

\bibitem[Van~Buuren \& Groothuis-Oudshoorn(2011)Van~Buuren and Groothuis-Oudshoorn]{van2011}
Van~Buuren, S. and Groothuis-Oudshoorn, K.
\newblock {mice: Multivariate imputation by chained equations in R}.
\newblock \emph{Journal of statistical software}, 45\penalty0 (1):\penalty0 1--67, 2011.

\bibitem[van~der Maaten \& Hinton(2008)van~der Maaten and Hinton]{maaten2008}
van~der Maaten, L. and Hinton, G.
\newblock {Visualizing Data using t-SNE}.
\newblock \emph{Journal of Machine Learning Research}, 9\penalty0 (86):\penalty0 2579--2605, 2008.

\bibitem[Vaswani et~al.(2017)Vaswani, Shazeer, Parmar, Uszkoreit, Jones, Gomez, Kaiser, and Polosukhin]{vaswani2017}
Vaswani, A., Shazeer, N., Parmar, N., Uszkoreit, J., Jones, L., Gomez, A.~N., Kaiser, L.~u., and Polosukhin, I.
\newblock {Attention is All you Need}.
\newblock In \emph{Advances in Neural Information Processing Systems}, volume~30, 2017.

\bibitem[Yadav et~al.(2018)Yadav, Steinbach, Kumar, and Simon]{yadav2018}
Yadav, P., Steinbach, M., Kumar, V., and Simon, G.
\newblock {Mining Electronic Health Records (EHRs): A Survey}.
\newblock \emph{ACM Comput. Surv.}, 50\penalty0 (6), 2018.

\bibitem[Yi et~al.(2020)Yi, Lee, Kim, Hwang, and Yang]{yi2020}
Yi, J., Lee, J., Kim, K.~J., Hwang, S.~J., and Yang, E.
\newblock {Why Not to Use Zero Imputation? Correcting Sparsity Bias in Training Neural Networks}.
\newblock In \emph{International Conference on Learning Representations}, 2020.

\bibitem[Yoon et~al.(2018)Yoon, Jordon, and van~der Schaar]{yoon2018}
Yoon, J., Jordon, J., and van~der Schaar, M.
\newblock {GAIN: Missing Data Imputation using Generative Adversarial Nets}.
\newblock In \emph{Proceedings of the 35th International Conference on Machine Learning}, volume~80 of \emph{Proceedings of Machine Learning Research}, pp.\  5689--5698. PMLR, 2018.

\bibitem[Yoon et~al.(2019)Yoon, Zame, and van~der Schaar]{yoon2019}
Yoon, J., Zame, W.~R., and van~der Schaar, M.
\newblock {Estimating Missing Data in Temporal Data Streams Using Multi-Directional Recurrent Neural Networks}.
\newblock \emph{IEEE Transactions on Biomedical Engineering}, 66\penalty0 (5):\penalty0 1477--1490, 2019.

\bibitem[Zhang et~al.(2017)Zhang, Zheng, and Qi]{zhang2017}
Zhang, J., Zheng, Y., and Qi, D.
\newblock {Deep Spatio-Temporal Residual Networks for Citywide Crowd Flows Prediction}.
\newblock In \emph{Proceedings of the AAAI Conference on Artificial Intelligence}, volume~31, 2017.

\end{thebibliography}

\clearpage
\appendix
\onecolumn

\section{Datasets Details}
\label{sec:app_datasets}

Three publicly available clinical datasets were used:

\emph{The PhysioNet Challenge 2012}: 
\url{https://physionet.org/content/challenge-2012/1.0.0/} \\[1mm]
\emph{MIMIC-\Romannum{3}}: 
\url{https://physionet.org/content/mimiciii/1.4/} \\[1mm]
\emph{The PhysioNet Challenge 2019}: 
\url{https://physionet.org/content/challenge-2019/1.0.0/} \\[1mm]

\subsection{Variables description}
Tables~\ref{tab:var_p12}, \ref{tab:var_m3}, and \ref{tab:var_p19} show 
all variables of PhysioNet 2012, MIMIC-\Romannum{3}, and PhysioNet 2019, 
respectively.

\begin{table}[h]
\caption{Variables list of \emph{The PhysioNet Challenge 2012}}
\label{tab:var_p12}
    \begin{center}
    \begin{small}
        \begin{tabular}{llccc}
            \toprule
            Index & Variable Name & Variable Type & Data Type & Missing Rate \\
            \midrule
            0 & Age & Static & Numerical & 0.000 \\
            1 & Gender & Static & Categorical & 0.075 \\
            2 & Height & Static & Numerical & 0.474 \\
            3 & ICUType & Static & Categorical & 0.000 \\
            4 & Albumin & Time-series & Numerical & 0.992 \\
            5 & ALP & Time-series & Numerical & 0.990 \\
            6 & ALT & Time-series & Numerical & 0.989 \\
            7 & AST & Time-series & Numerical & 0.989 \\
            8 & Bilirubin & Time-series & Numerical & 0.989 \\
            9 & BUN & Time-series & Numerical & 0.954 \\
            10 & Cholesterol & Time-series & Numerical & 0.999 \\
            11 & Creatinine & Time-series & Numerical & 0.953 \\
            12 & DiasABP & Time-series & Numerical & 0.514 \\
            13 & FiO2 & Time-series & Numerical & 0.892 \\
            14 & GCS & Time-series & Numerical & 0.794 \\
            15 & Glucose & Time-series & Numerical & 0.957 \\
            16 & HCO3 & Time-series & Numerical & 0.955 \\
            17 & HCT & Time-series & Numerical & 0.939 \\
            18 & HR & Time-series & Numerical & 0.236 \\
            19 & K & Time-series & Numerical & 0.952 \\
            20 & Lactagte & Time-series & Numerical & 0.973 \\
            21 & Mg & Time-series & Numerical & 0.955 \\
            22 & MAP & Time-series & Numerical & 0.519 \\
            23 & MechVent & Time-series & Categorical & 0.896 \\
            24 & Na & Time-series & Numerical & 0.955 \\
            25 & NIDiasABP & Time-series & Numerical & 0.672 \\
            26 & NIMAP & Time-series & Numerical & 0.676 \\
            27 & NISysABP & Time-series & Numerical & 0.671 \\
            28 & PaCO2 & Time-series & Numerical & 0.922 \\
            29 & PaO2 & Time-series & Numerical & 0.922 \\
            30 & pH & Time-series & Numerical & 0.919 \\
            31 & Platelets & Time-series & Numerical & 0.953 \\
            32 & RespRate & Time-series & Numerical & 0.816 \\
            33 & SaO2 & Time-series & Numerical & 0.973 \\
            34 & SysABP & Time-series & Numerical & 0.513 \\
            35 & Temp & Time-series & Numerical & 0.712 \\
            36 & TroponinI & Time-series & Numerical & 0.999 \\
            37 & TroponinT & Time-series & Numerical & 0.993 \\
            38 & Urine & Time-series & Numerical & 0.548 \\
            39 & WBC & Time-series & Numerical & 0.957 \\
            40 & Weight & Time-series & Numerical & 0.569 \\
            \bottomrule
        \end{tabular}
    \end{small}
    \end{center}
\end{table}

\begin{table}[h]
\caption{Variables list of \emph{MIMIC- \Romannum{3}}}
\label{tab:var_m3}
    \begin{center}
    \begin{small}
        \begin{tabular}{llccc}
            \toprule
            Index & Variable Name & Variable Type & Data Type & Missing Rate \\
            \midrule
            0 & Age & Static & Numerical & 0.000 \\
            1 & Gender & Static & Categorical & 0.000 \\
            2 & DBP & Time-series & Numerical & 0.242 \\
            3 & FiO2 & Time-series & Numerical & 0.958 \\
            4 & Glucose & Time-series & Numerical & 0.825 \\
            5 & HR & Time-series & Numerical & 0.229 \\
            6 & pH & Time-series & Numerical & 0.924 \\
            7 & RR & Time-series & Numerical & 0.225 \\
            8 & SBP & Time-series & Numerical & 0.242 \\
            9 & SpO2 & Time-series & Numerical & 0.248 \\
            10 & Temp(C) & Time-series & Numerical & 0.909 \\
            11 & Temp(F) & Time-series & Numerical & 0.839 \\
            12 & TGCS & Time-series & Numerical & 0.877 \\
            \bottomrule
        \end{tabular}
    \end{small}
    \end{center}
\end{table}

\begin{table}[h]
\caption{Variables list of \emph{The PhysioNet Challenge 2019}}
\label{tab:var_p19}
    \begin{center}
    \begin{small}
        \begin{tabular}{llccc}
            \toprule
            Index & Variable Name & Variable Type & Data Type & Missing Rate \\
            \midrule
            0 & Age & Static & Numerical & 0.000 \\
            1 & Gender & Static & Categorical & 0.000 \\
            2 & Unit1 & Static & Categorical & 0.390 \\
            3 & Unit2 & Static & Categorical & 0.390 \\
            4 & HospAdmTime & Static & Numerical & 0.000 \\
            5 & HR & Time-series & Numerical & 0.100 \\
            6 & O2Sat & Time-series & Numerical & 0.132 \\
            7 & Temp & Time-series & Numerical & 0.662 \\
            8 & SBP & Time-series & Numerical & 0.146 \\
            9 & MAP & Time-series & Numerical & 0.126 \\
            10 & DBP & Time-series & Numerical & 0.316 \\
            11 & Resp & Time-series & Numerical & 0.155 \\
            12 & EtCO2 & Time-series & Numerical & 0.963 \\
            13 & BaseExcess & Time-series & Numerical & 0.946 \\
            14 & HCO3 & Time-series & Numerical & 0.958 \\
            15 & FiO2 & Time-series & Numerical & 0.916 \\
            16 & pH & Time-series & Numerical & 0.931 \\
            17 & PaCO2 & Time-series & Numerical & 0.944 \\
            18 & SaO2 & Time-series & Numerical & 0.965 \\
            19 & AST & Time-series & Numerical & 0.984 \\
            20 & BUN & Time-series & Numerical & 0.931 \\
            21 & Alkalinephos & Time-series & Numerical & 0.984 \\
            22 & Calcium & Time-series & Numerical & 0.941 \\
            23 & Chloride & Time-series & Numerical & 0.955 \\
            24 & Creatinine & Time-series & Numerical & 0.939 \\
            25 & Bilirubin direct & Time-series & Numerical & 0.998 \\
            26 & Glucose & Time-series & Numerical & 0.829 \\
            27 & Lactate & Time-series & Numerical & 0.973 \\
            28 & Magnesium & Time-series & Numerical & 0.937 \\
            29 & Phosphate & Time-series & Numerical & 0.960 \\
            30 & Potassium & Time-series & Numerical & 0.907 \\
            31 & Bilirubin total & Time-series & Numerical & 0.985 \\
            32 & TropinI & Time-series & Numerical & 0.990 \\
            33 & Hct & Time-series & Numerical & 0.911 \\
            34 & Hgb & Time-series & Numerical & 0.926 \\
            35 & PTT & Time-series & Numerical & 0.971 \\
            36 & WBC & Time-series & Numerical & 0.936 \\
            37 & Fibrinogen & Time-series & Numerical & 0.994 \\
            38 & Platelets & Time-series & Numerical & 0.941 \\
            \bottomrule
        \end{tabular}
    \end{small}
    \end{center}
\end{table}

\clearpage
When admitted to ICU, various variables are collected with multiple observations 
to monitor the status of a patient. Demographics variables are collected when the 
patient is admitted and physiological variables such as vital signs and blood test 
are measured as needed after admission. This causes a irregularly sampled time series 
with different variable types. The missing rates in Tables~\ref{tab:var_p12}, 
\ref{tab:var_m3}, and \ref{tab:var_p19} were measured by the conventional structured 
representation of time series. All missing rates were measured on the training dataset.

\subsection{Dataset preprocessing}
For the three datasets, we converted categorical data to one-hot-encoding 
representation. Additionally, we normalized all numerical data to follow the 
standard Gaussian distribution for the stability of the learning.

\subsubsection{The PhysioNet Challenge 2012}
We used all variables and all observations without discretizing data into fixed hourly 
bins. 

\subsubsection{MIMIC-\Romannum{3}}
The dataset is a large relational EHR database containing several tables. Thus, we 
extracted the necessary data based on the criteria in \emph{The PhysioNet Challenge 
2012} \cite{silva2012}, the criteria used in a benchmark method \cite{purushotham2018}, 
and IP-Nets \cite{shukla2019}. 

\subsubsection{The PhysioNet Challenge 2019}
Because this dataset was already provided as the conventional structured representation 
with discretizing into fixed hourly bins, we used the dataset as it is.

\section{Training Details}
\label{sec:app_training}

\subsection{Experiment settings}
We randomly split three datasets into training, validation, and test datasets with 
a 6:2:2 ratio keeping the class label ratio, except for \emph{The PhysioNet 2012} 
dataset, which is described in Section~\ref{sec:datasets}. Training process was 
stopped when there was no improvement of AUROC on the validation dataset. The 
hyperparamters of all models were selected based on the best performance on the 
validation dataset. All detailed descriptions of the hyperparameters are described 
in Table~\ref{tab:hp_search}. For the mTAND models, we followed the best hyperparameter 
used in the original paper \cite{shukla2021}. For the TDE models, because the number 
of variables is different for each dataset, the hyperparameter search range was set 
slightly different. With the optimal parameters of the best performance on the 
validation dataset, we measured the statistical performance on the test dataset 
using the bootstrap method \cite{james2013}.

\begin{table}[h]
\caption{Hyperparameter search}
\label{tab:hp_search}
    \begin{center}
    \begin{adjustbox}{width=0.8\linewidth}
    \begin{small}
        \begin{tabular}{lccccc}
            \toprule
            Model & Hyperparameter Name & Search Range & PhysioNet 2012 & MIMIC-\Romannum{3} & PhysioNet 2019 \\
            \midrule
            {All} & Batch size & - & 256 & 256 & 256 \\
            {} & Optimizer & - & Adam & Adam & Adam \\
            {} & Learning rate & - & 0.001 & 0.001 & 0.001 \\
            \midrule
            RNN & Dropout rate & (0.0, 0.2, 0.5) & 0.5 & 0.0 & 0.0 \\
            {} & Recurrent hidden unit & (32, 64, 128, 256, 512) & 32 & 32 & 128 \\[3mm]
            LSTM & Dropout rate & (0.0, 0.2, 0.5) & 0.2 & 0.5 & 0.2 \\
            {} & Recurrent hidden unit & (32, 64, 128, 256, 512) & 128 & 64 & 256 \\[3mm]
            GRU & Dropout rate & (0.0, 0.2, 0.5) & 0.5 & 0.2 & 0.0 \\
            {} & Recurrent hidden unit & (32, 64, 128, 256, 512) & 128 & 128 & 64 \\[3mm]
            GRU-D & Dropout rate & (0.0, 0.2, 0.5) & 0.2 & 0.2 & 0.0 \\
            {} & Recurrent hidden unit & (32, 64, 128, 256, 512) & 64 & 32 & 256 \\[3mm]
            BRITS & Dropout rate & (0.0, 0.2, 0.5) & 0.0 & 0.0 & {-} \\
            {} & Recurrent hidden unit & (32, 64, 128, 256, 512) & 64 & 64 & {-} \\
            {} & Loss weight for imp & (0.0, 0.3, 0.5, 0.8, 1.0) & 0.0 & 1.0 & {-} \\[3mm]
            mTAND-Enc & Reference number & {-} & 128 & 128 & {-} \\
            {} & Time embedding & {-} & 128 & 128 & {-} \\
            {} & Multi-head number & {-} & 1 & 1 & {-} \\
            {} & Q, K, V dimension & {-} & 128 & 128 & {-} \\
            {} & Recurrent hidden unit & {-} & 256 & 128 & {-} \\
            {} & FFN unit & {-} & 50 & 50 & {-} \\
            {} & Latent dimension & {-} & 20 & 20 & {-} \\
            {} & Classifier unit & {-} & 300 & 300 & {-} \\[3mm]
            mTAND-Full & Loss weight for pred & {-} & 100 & 5 & {-} \\
            {} & Loss weight for recon & {-} & 1 & 1 & {-} \\
            {} & Reference number & {-} & 128 & 128 & {-} \\
            {} & Time embedding & {-} & 128 & 128 & {-} \\
            {} & Multi-head number & {-} & 1 & 1 & {-} \\
            {} & Q, K, V dimension & {-} & 128 & 128 & {-} \\
            {} & Recurrent hidden unit - enc & {-} & 256 & 256 & {-} \\
            {} & Recurrent hidden unit - dec & {-} & 50 & 50 & {-} \\
            {} & FFN unit & {-} & 50 & 50 & {-} \\
            {} & Latent dimension & {-} & 20 & 128 & {-} \\
            {} & Classifier unit & {-} & 300 & 300 & {-} \\
            \midrule
            TDE-Mean & Dropout rate & (0.0, 0.2, 0.5) & 0.2 & 0.2 & 0.2 \\
            {} & Static embedding & (2, 4, 6) & 4 & 2 & 2 \\
            {} & Time-series embedding & (8, 16, 32, 64) & 16 & 64 & 32 \\
            {} & Time embedding & (8, 16, 32, 64) & 32 & 8 & 8 \\
            {} & Aggregation embedding & (16, 32, 64) & 16 & 16 & 64 \\
            {} & Recurrent hidden unit & (32, 64, 128) & 128 & 16 & 64 \\[3mm]
            TDE-Attn & Dropout rate & (0.0, 0.2, 0.5) & 0.5 & 0.2 & 0.2 \\
            {} & Static embedding & (2, 4) & 2 & 2 & 2 \\
            {} & Time-series embedding & (8, 16, 32, 64) & 16 & 128 & 16 \\
            {} & Q, K, V dimension & (4, 8, 16, 32) & 8 & 16 & 4 \\
            {} & Multi-head number & (1, 2, 4, 8) & 1 & 1 & 8 \\
            {} & Aggregation embedding & (8, 16, 32, 64) & 8 & 16 & 32 \\
            {} & Recurrent hidden unit & (32, 64, 128, 256) & 256 & 64 & 256 \\
            \bottomrule
        \end{tabular}
    \end{small}
    \end{adjustbox}
    \end{center}
\end{table}

\subsection{Computing environment}
All experiments were conducted under the same environments of NVIDIA Geforce RTX 2080 Ti.

\section{Ablation Study}
\label{sec:app_ablation}

We performed ablation experiments to identify the effect of softmax and non-softmax 
attention score in Equation~(\ref{eq:attn}) for classification performance. 
Table~\ref{tab:ablation} compares the performance on PhysioNet 2012, MIMIC-\Romannum{3}, 
and PhysioNet 2019.

\begin{table}[h]
\caption{Performance comparison for attention method}
\label{tab:ablation}
    \begin{center}
    \begin{small}
        \begin{tabular}{lccc}
            \toprule
            Dataset & Attention method & AUROC & AUPRC \\
            \midrule
            PhysioNet 2012      & Softmax       & {0.808 $\pm$ 0.008} & {0.430 $\pm$ 0.021} \\
            {}                  & Non-softmax   & {\textbf{0.848 $\pm$ 0.009}} & {\textbf{0.532 $\pm$ 0.022}} \\[3mm]
            MIMIC-\Romannum{3}  & Softmax       & {0.811 $\pm$ 0.008} & {0.446 $\pm$ 0.020} \\
            {}                  & Non-softmax   & {\textbf{0.818 $\pm$ 0.008}} & {\textbf{0.475 $\pm$ 0.019}} \\[3mm]
            PhysioNet 2019      & Softmax       & {0.815 $\pm$ 0.007} & {0.088 $\pm$ 0.006} \\
            {}                  & Non-softmax   & {\textbf{0.819 $\pm$ 0.007}} & {\textbf{0.095 $\pm$ 0.008}} \\
            \bottomrule
        \end{tabular}
    \end{small}
    \end{center}
\end{table}

\end{document}